\pdfoutput=1
\documentclass[conference]{IEEEtran}
\usepackage{times}
\usepackage[utf8]{inputenc}
\usepackage[T1]{fontenc}
\usepackage{cite}
\usepackage{amsmath,amssymb}
\usepackage{graphicx}
\usepackage{booktabs}
\usepackage{url}
\usepackage{hyperref}
\hypersetup{colorlinks=true, linkcolor=black, citecolor=black, urlcolor=blue}

\begin{document}

\title{Nori A3: A Bimanual Mobile Manipulator\\at the Appliance Price Point}

\author{\IEEEauthorblockN{Antonio Li}
\IEEEauthorblockA{Nori Robotics, San Francisco, CA\\
antoniositongli@norirobotics.com}}

\maketitle

\begin{abstract}
We present Nori A3, a 19-DoF bimanual mobile manipulator that ships
assembled for \$1{,}688. In a 45\,$\times$\,45\,cm footprint at
20.4\,kg, a three-stage telescoping column carries the head and both
arms from 69\,cm to 145\,cm in height, giving floor-to-counter reach;
each 55\,cm arm has 7+1 DoF and a 1.5\,kg payload, alongside lidar
navigation and four cameras. This is roughly a fifth of the parts cost
of the cheapest comparable research platform. The capabilities usually
lost at this price are recovered in software rather than by buying
more expensive parts. A protection architecture eliminates the thermal
burn-out of commodity serial-bus servos, attacking both the stall
conditions that cause it and, through a two-tier interlock, the
temperature rise itself. A sensorless grip-force channel recovers a
continuous force estimate from the servo's own current register.
A3 has been shipping to customers since July 2026; we give the design
rationale and the per-joint actuator selection.
\end{abstract}

\section{Introduction}
\label{sec:intro}

Capable mobile manipulators have stalled at a parts-cost floor near
\$10{,}000, with YOR~\cite{anjaria2026yor} at a \$9{,}250 bill of
materials and TidyBot++~\cite{wu2024tidybotpp} at \$5{,}000--6{,}000
for a single arm (Table~\ref{tab:cost}). Cheaper platforms exist on
commodity serial-bus servos, but the literature is consistent that the
tier gives up payload, reach, and durability; YOR names all three in
describing XLeRobot~\cite{wang2025xlerobot}.

We attribute those limits to execution rather than to anything
inherent in the price tier. Reach follows from mechanical layout and
payload from transmission and mounting geometry, so both are open to
ordinary design effort, while durability, the objection raised most
often, reduces to one thermal failure mode that can be prevented in
software.

\begin{figure*}[t]
\centering
\includegraphics[height=0.38\textwidth]{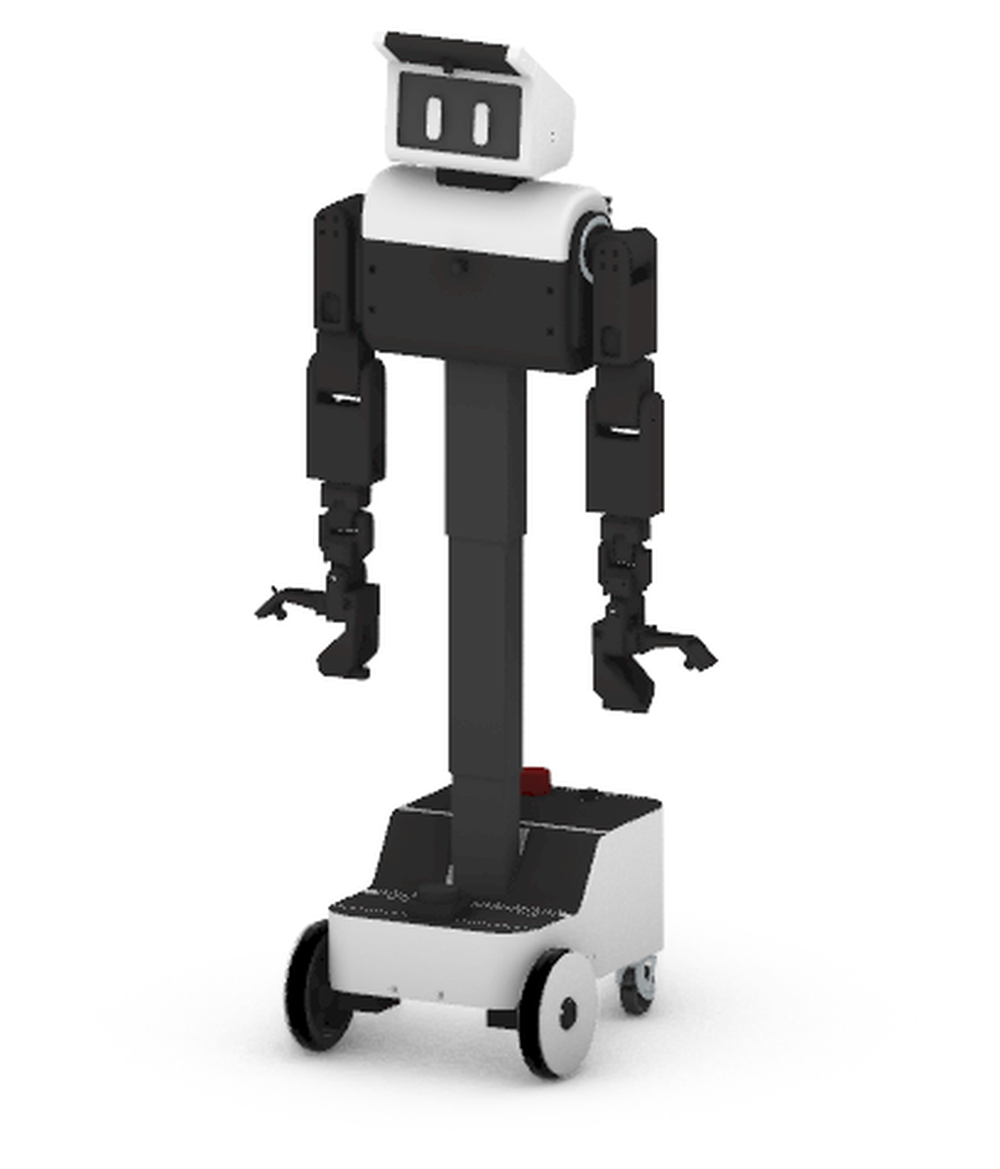}\hspace{16pt}%
\includegraphics[height=0.38\textwidth]{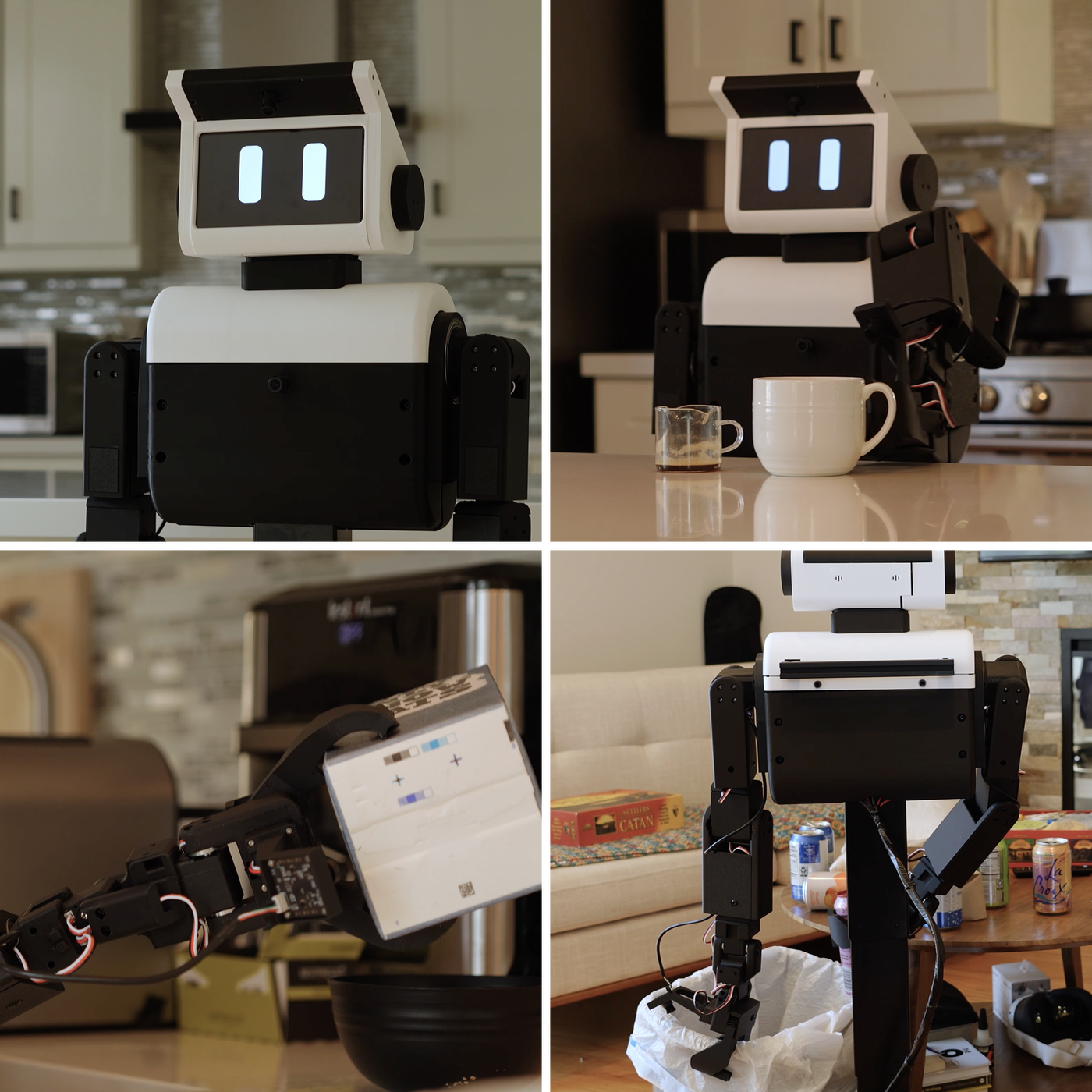}
\caption{Nori A3, assembled retail price \$1{,}688. Left: column
extended, 19 actuated DoF in a 45\,$\times$\,45\,cm footprint at
20.4\,kg. Right: operating in unmodified homes.}
\label{fig:robot}
\end{figure*}

Nori A3 (Fig.~\ref{fig:robot}) is a 19-DoF bimanual mobile
manipulator that ships assembled for \$1{,}688 and gives up none of
the three. The price comparison runs in our favor for a reason worth
stating: \$9{,}250 for YOR is a bill of materials a lab must source
and assemble, whereas \$1{,}688 is a retail price covering assembly,
warranty, and support. A3 costs less than the parts of a comparable
platform, requires none of the build labor, and delivers the reach,
bimanual workspace, and sensing of machines costing five times as
much. It is also a product, and we do not publish the full mechanical
package. What we do instead is expose the kinematics in a browser and
the end effector on an open connector (Sec.~\ref{sec:ops}), and
open-source the actuator-protection layer, which transfers to other
platforms.

Contributions:

\begin{enumerate}
    \item \textbf{A form factor for bimanual mobile manipulation at
    the appliance price point.} A wheeled base with a telescoping
    column carrying head and both arms gives floor-to-counter reach
    without legged complexity and collapses to work under counters, at
    roughly a fifth of the parts price of the cheapest comparable
    research platform.
    \item \textbf{Software that makes commodity servos viable.} A
    layered protection architecture eliminates the thermal failure
    that dominates durability complaints about this actuator class,
    acting on its cause and, through a two-tier interlock, on its
    effect. A sensorless grip-force channel recovers a continuous
    force signal from the current register those servos already
    expose. Neither adds hardware cost.
    \item \textbf{Design rationale} for a platform that has been
    shipping to customers since July 2026, including the reasoning
    behind per-joint actuator selection.
\end{enumerate}

\section{Related Work}
\label{sec:related}

\paragraph{Low-cost mobile manipulators}
Hello Robot Stretch~\cite{kemp2022stretch} established the compact
single-arm form factor for indoor environments and is the closest
commercial analogue to this work, a closed product justified by design
rationale rather than released CAD. Mobile
ALOHA~\cite{fu2024mobilealoha} demonstrated bimanual whole-body
teleoperation, and TidyBot++~\cite{wu2024tidybotpp} cut cost with a
holonomic base and vertical lift but a single arm.
YOR~\cite{anjaria2026yor} is the closest recent comparison, pairing an
omnidirectional swerve base and a telescopic column with two 6-DoF
compliant arms. We agree with its central conclusion, that a wheeled
base with a vertical lift and two arms is the right shape for indoor
mobile manipulation, and differ in cost tier and in being a
manufactured product. An earlier version of this
platform~\cite{li2026noribot} was built on the
XLeRobot~\cite{wang2025xlerobot} base; A3 shares none of that
hardware, and AhaRobot~\cite{cui2025aharobot} occupies the same low
tier.

\paragraph{Force sensing without a force sensor}
Estimating contact force from motor current is standard in industrial
robotics, and Universal Robots and Robotiq ship sensorless force
estimation on cobots and parallel grippers. Research alternatives add
hardware: embedded air channels~\cite{liu2025forte}, external
cameras~\cite{collins2022vfts}, discrete tactile
arrays~\cite{xu2023tacfr}. Several authors note that current alone is
a weak proxy on a rigid finger~\cite{collins2022vfts}; we agree, and
argue in Sec.~\ref{sec:force} that it becomes continuous and useful
once the finger is compliant. Our protections are closest in spirit to
bio-inspired grasping controllers~\cite{lach2023bioinspired} that
detect contact through phase transitions, without a dedicated sensor.

\paragraph{Why unit cost bounds data collection}
$\pi_0$~\cite{black2024pi0} and Open
X-Embodiment~\cite{openxembodiment2024} establish that pooling
demonstrations across robots improves generalist policies, and the
tooling from teleoperation to policy has standardized around ACT-style
architectures~\cite{zhao2023act} and open
pipelines~\cite{cadene2024lerobot}, shifting the binding constraint
toward the number of robots available to demonstrate on. This is a
hardware paper and does not address the policy side, but the point
gives unit cost a significance beyond affordability.

\section{Hardware Design}
\label{sec:hardware}

\subsection{Design principles}

\begin{itemize}
    \item \emph{Protect the actuators rather than upgrade them.}
    Engineering effort went into Sec.~\ref{sec:actuators} instead of
    budget going into better motors.
    \item \emph{Vertical reach over locomotion complexity.} A
    telescoping column reaches floor to counter without balance
    control or legged actuation.
    \item \emph{One actuator family, graded by joint load.} Three
    torque grades of Feetech STS-series servo, matched to the moment
    each joint carries (Table~\ref{tab:actuators}).
    \item \emph{Manufacturability from day one.} Part count and
    assembly time were design criteria alongside performance; A3 is
    assembled in San Francisco in production batches.
\end{itemize}

\subsection{Platform}

A3 occupies a 45\,$\times$\,45\,cm footprint at 20.4\,kg, light
enough for one person to carry between rooms or into a vehicle. It has
19 actuated DoF: 7+1 per arm, the lift, and two differential drive
wheels, with passive casters taking the remaining ground contact.

\begin{figure}[!tb]
\centering
\includegraphics[width=0.88\columnwidth]{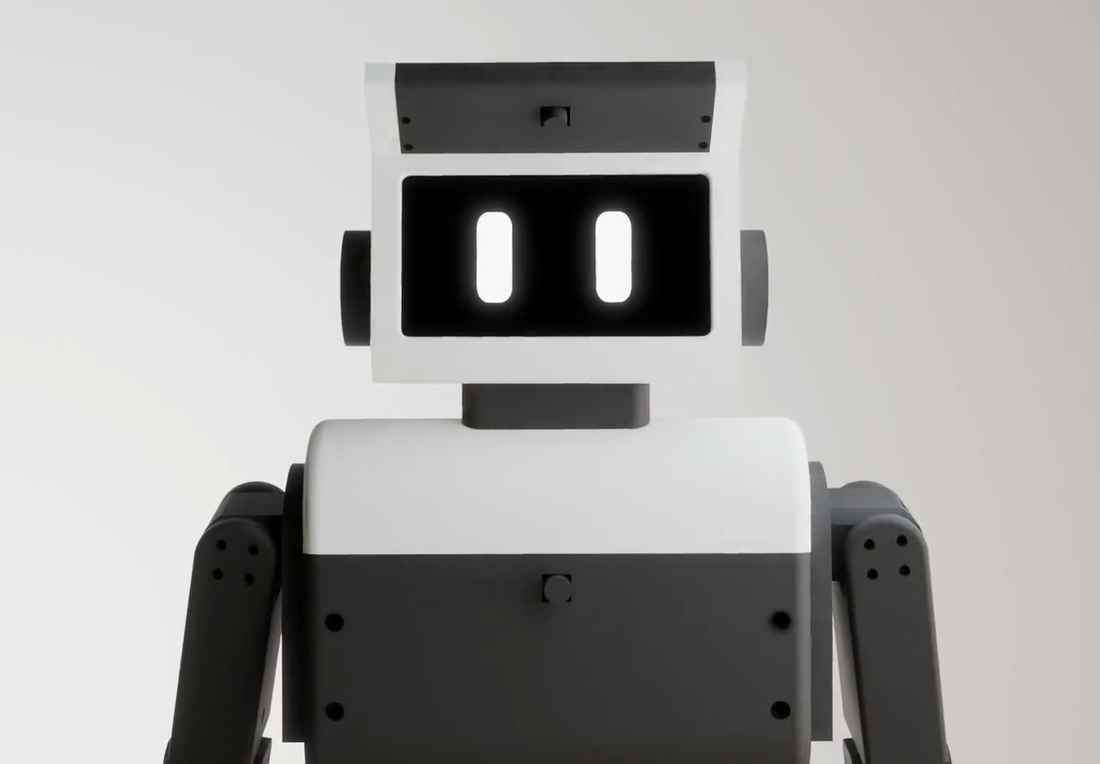}
\caption{Head and torso, front view.}
\label{fig:front}
\end{figure}

Perception is four 720p RGB cameras at up to 30\,fps on the grippers,
head, and neck (Fig.~\ref{fig:front}), plus a lidar with 12\,m range,
8--12\,Hz scan rate, and 0.72$^\circ$ angular resolution at 10\,Hz. A
speaker and microphone support spoken commands. Onboard power provides
6--8 hours of operation. The cylindrical elements flanking the display
are cosmetic and tapped M3, usable as mounting points for additional
sensors.

\begin{figure}[!tb]
\centering
\includegraphics[width=0.88\columnwidth]{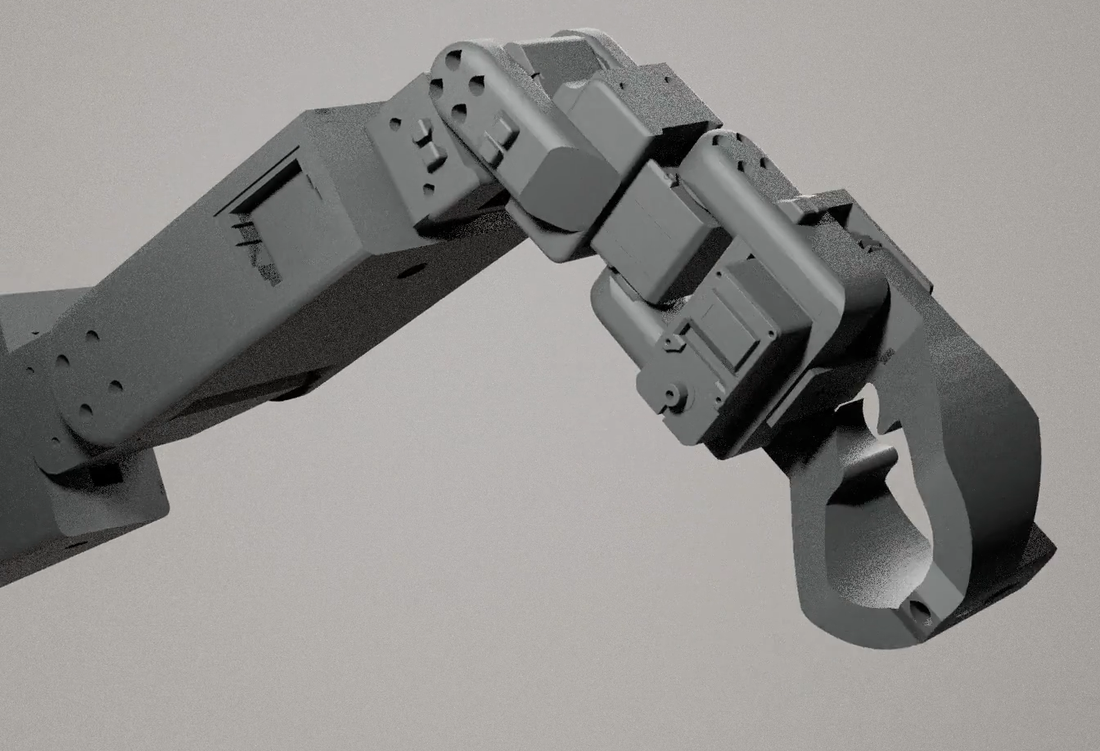}
\caption{Forearm, wrist, and gripper.}
\label{fig:arm}
\end{figure}

\paragraph{Arms}
Each arm has seven actuated joints plus a gripper
(Fig.~\ref{fig:arm}), reaching 55\,cm at a 1.5\,kg payload. The
seventh joint acts about the wrist axis, letting the end effector hold
a pose while the elbow reconfigures around obstacles, the common case
when reaching into a cabinet or across a crowded counter.
Against YOR's 6-DoF PiPER arms at \$2{,}500 each, that redundancy
costs a fraction as much because the actuators are commodity
serial-bus servos rather than industrial modules;
Sec.~\ref{sec:actuators} argues why the substitution is viable. The
arms terminate in open connector joints for custom end effectors.

\begin{figure}[!tb]
\centering
\includegraphics[width=0.88\columnwidth]{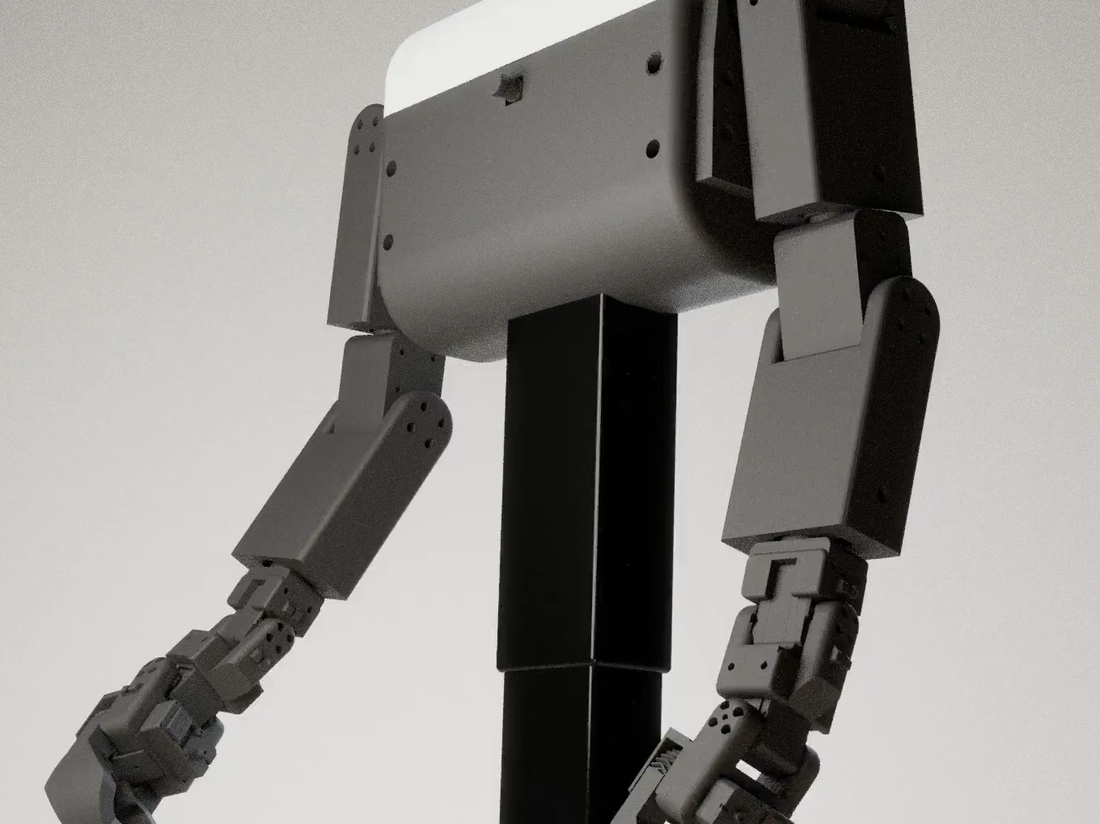}
\caption{Shoulder block and upper lift stage.}
\label{fig:torso}
\end{figure}

\paragraph{Lift}
A three-stage steel telescoping column, of the type used for
height-adjustable desks and comparable to YOR's, replaces the
fixed-stroke ball screw of the prior version. Nesting stages decouple
stowed height from extended reach, where a fixed-length screw would
have had to stand as tall as its full travel at all times. A single
electric motor drives 76\,cm of travel at 30\,mm/s, supporting a
55\,kg vertical load.

Lateral stiffness rather than lifting capacity governs the design.
The column carries the head and both arms, so rotational play at the
carriage reaches the end effector amplified by arm length, and does so
differently at different extensions. In the prior version an early
single-rod prototype had enough play under arm moment load that
policies trained at one height degraded at another. Both arms mount to
one cross-member carried by the column (Fig.~\ref{fig:torso}), which
makes the arm-to-arm transform invariant to height, so the bimanual
workspace translates as a single body. Carriage deviation under rated
arm load, measured across the column's extension range, is the figure
that would quantify this, and we report it in a future revision.

\paragraph{Base}
The base is differentially driven rather than holonomic, an explicit
cost decision: YOR spends \$2{,}700 of its \$9{,}250 on four swerve
modules, more than A3's entire retail price. Lateral repositioning is
therefore a compound maneuver rather than a strafe. The
45\,$\times$\,45\,cm footprint keeps the turning circle small enough
for domestic interiors.

\subsection{Actuation and compute}

\begin{table}[t]
\centering
\caption{Actuator placement, one arm. Both arms are identical.}
\label{tab:actuators}
\footnotesize
\begin{tabular}{llll}
\toprule
\textbf{\#} & \textbf{Joint} & \textbf{Servo} & \textbf{Stall (kg$\cdot$cm)} \\
\midrule
1 & Shoulder, first axis  & STS3095 & 95 \\
2 & Shoulder, second axis & STS3095 & 95 \\
3 & Bicep rotation        & STS3250 & 50 \\
4 & Bicep                 & STS3095 & 95 \\
5 & Forearm roll          & STS3215 & 30 \\
6 & Wrist, first axis     & STS3250 & 50 \\
7 & Wrist, second axis    & STS3215 & 30 \\
8 & Gripper               & STS3215 & 30 \\
\bottomrule
\end{tabular}
\end{table}

Table~\ref{tab:actuators} gives the per-joint selection. Grade
follows the gravitational moment a joint holds rather than its
distance from the base, which sorts the axes into two classes. Pitch axes carry the weight of everything distal to them
and take the higher grades: both shoulder axes and the bicep at
95\,kg$\cdot$cm, the wrist pitch at 50. Roll axes rotate the limb
about its own long axis, hold almost no gravitational moment, and take
one grade below the pitch axis immediately proximal: bicep rotation at
50 under the 95 above it, forearm roll at 30 under the 95 at the
bicep. Sizing every joint to the shoulder requirement would have put
95\,kg$\cdot$cm actuators at the wrist, adding mass exactly where it
costs most, since distal mass loads every joint upstream.

All three grades share a protocol, a register map, and a wiring
standard, so calibration, the bus driver, stall detection, thermal
supervision, and force estimation are written once and apply to every
joint. Grading the hardware per joint therefore costs nothing in
software complexity.

Onboard compute is a Raspberry Pi 5 with 4\,GB of RAM, handling bus
I/O, the control loop, the protection tiers of
Sec.~\ref{sec:actuators}, and sensor acquisition. Inference runs
off-board on a backend serving language-model control and VLA or ACT
policies. This keeps per-unit compute at the price of a single-board
computer and lets a policy improvement reach deployed units without a
hardware change.

\subsection{Cost}

\begin{table}[t]
\centering
\caption{Cost against comparable platforms. We do not normalize the
bases, since the difference is itself informative: a lab acquiring YOR
pays \$9{,}250 and then supplies the build, while a lab acquiring A3
pays \$1{,}688 and unpacks it.}
\label{tab:cost}
\footnotesize
\begin{tabular}{llr}
\toprule
\textbf{Platform} & \textbf{Basis} & \textbf{Cost (USD)} \\
\midrule
Mobile ALOHA~\cite{fu2024mobilealoha}      & BOM    & 32{,}000 \\
Hello Robot Stretch~\cite{kemp2022stretch} & retail & $\sim$25{,}000 \\
YOR~\cite{anjaria2026yor}                  & BOM    & 9{,}250 \\
TidyBot++~\cite{wu2024tidybotpp}           & BOM    & 5{,}000--6{,}000 \\
AhaRobot~\cite{cui2025aharobot}            & BOM    & 1{,}000--2{,}000 \\
\textbf{Nori A3 (ours)}                    & retail & \textbf{1{,}688} \\
XLeRobot~\cite{wang2025xlerobot}           & BOM    & 660 \\
\bottomrule
\end{tabular}
\end{table}

\section{Making Commodity Actuators Viable}
\label{sec:actuators}

Feetech STS-class servos fail along a deterministic chain that
terminates in heat: a goal position is commanded outside the joint
range or against an obstacle, the integral term ramps, the motor draws
stall current, and sustained current raises winding temperature until
the motor is damaged. The failure is thermal throughout, not
mechanical, which dictates a defense at two points.
Sec.~\ref{sec:protection} prevents the conditions producing sustained
current; Sec.~\ref{sec:thermal} monitors temperature directly. Neither
suffices alone, since cause-side protection cannot cover legitimate
high-load operation sustained too long, and temperature monitoring on
its own would let the robot repeatedly approach its limit in normal
use. To our knowledge no published low-cost platform documents either
layer.

\subsection{Preventing stall}
\label{sec:protection}

\paragraph{Calibration clamping} Stock LeRobot's
\texttt{\_unnormalize()} did not clamp goal positions to the
calibrated range on \texttt{DEGREES}-mode joints. We clamp at the bus
layer, so no commanded position can drive a joint into a hard stop.

\paragraph{Stall detection} Each control loop we read
\texttt{Present\_Current} (register 69) alongside
\texttt{Present\_Position} and the goal, maintain a $k=15$-loop
window, and declare a stall when current is high but position is
static across the window. On stall the gripper drops to a reduced
grip torque, holding the object gently rather than crushing it; other
joints retry once at full torque, then drop to a reduced torque until
a new target is commanded.

\paragraph{Firmware backstops} Three EEPROM registers are written once
per motor as a fallback if clamp and detector both fail:
\texttt{Protection\_Current}, \texttt{Over\_Current\_Protection\_Time},
and \texttt{Max\_Torque\_Limit}, the last so the otherwise-volatile
SRAM torque limit survives a power cycle.

\subsection{Two-tier thermal interlock}
\label{sec:thermal}

We monitor temperature at two tiers, following the standard pattern
of an independent hardware interlock backing a software supervisor.

\paragraph{Hardware tier} Each motor carries an over-temperature
cutoff acting on that motor alone, disabling output at
$T_{\text{hw}} = 60^{\circ}$C, where damage begins. It is deliberately
simple, with no staging and no dependence on the bus being alive or
the host responsive, and it is not reconfigurable at runtime.

\paragraph{Software tier} The control loop polls per-motor temperature
alongside the current and position reads it already performs, and
tracks the trajectory rather than the instantaneous value. Warnings
escalate as any motor approaches $T_{\text{sw}} = 59^{\circ}$C,
letting an operator or scheduler reduce duty cycle before anything
trips. At $T_{\text{sw}}$ the supervisor disables the entire motor bus
rather than the offending motor alone.

The bus-wide response follows from the daisy-chain topology. Motors on
a shared chain sit in close proximity, and those still holding position
keep drawing current and dissipating heat into the same structure.
Cutting only the motor that tripped leaves its neighbours heating the
assembly it sits in, and leaves the arm in a configuration the
remaining motors must hold against gravity, raising their own draw.
Disabling the bus removes the heat source at once.

The $1^{\circ}$C offset is not thermal headroom; it exists to order
the two shutdowns and avoid a race. They leave the machine in
different states, a software trip being a clean, logged, bus-wide
shutdown the system recovers from on its own, while a hardware trip
drops one motor without announcement and the control loop must infer
what happened. Placing the supervisor one degree lower keeps the
recoverable path as the normal outcome.

\subsection{Sensorless grip force}
\label{sec:force}

We recover a continuous grip-force estimate from the same
\texttt{Present\_Current} register the stall detector already reads
($\sim$6.5\,mA per raw unit), paired with a soft TPU finger. The
compliance is what makes the signal useful. On a rigid finger, motor
current produces a stall-like step hard to disentangle from inertial
transients~\cite{collins2022vfts}; on the TPU finger, commanded travel
after contact deforms the finger rather than the object, so the
encoder continues to move slightly while current rises monotonically
with squeeze depth. The result is a continuous current-vs-squeeze ramp
instead of a discontinuous step.

We map raw current $I_t$ to a normalized estimate
\[
\hat{f}_t = \mathrm{clip}\!\left(
\frac{|I_t| - I_{\text{idle}}}{I_{\text{max}} - I_{\text{idle}}},\ 0,\ 1\right)
\]
with $I_{\text{idle}}$ a friction floor and $I_{\text{max}}$ the
empirical peak before the stall detector engages. Because the detector
caps current, the torque reduction acts as a saturation ceiling:
$\hat{f}_t = 1$ corresponds unambiguously to fully gripping. Constants
are calibrated per gripper.

The force read piggybacks on the stall detector's existing current
sync-read, since a second read would have doubled bus I/O. Every
demonstration records the signal as two observation channels, raw
current and normalized $\hat{f}_t$, while the action vector stays
joint positions only. Against dedicated tactile
hardware~\cite{xu2023tacfr,liu2025forte} we recover one scalar per
gripper rather than a contact distribution, which is the cost of using
a signal that is already on the bus.

\section{Operation and Deployment}
\label{sec:ops}

Operation and demonstration recording run through Nori Lab, a
browser-based interface at
\url{https://lab.norirobotics.com/nori/model}. Because it runs in the
browser rather than as a local install, the kinematics and workspace
can be driven and inspected without hardware. Recorded demonstrations
land in a LeRobot-compatible dataset
format~\cite{cadene2024lerobot}, so existing training pipelines apply
without conversion.

The first A3 units shipped to customers on July 21, 2026 and have
been produced in batches since. The protections of
Sec.~\ref{sec:actuators} were designed for a robot running
unsupervised in someone's home, with no engineer present to notice a
motor running hot, a stricter requirement than a lab platform faces
and the reason the hardware tier is independent of everything above
it.

\section{Scope and Tradeoffs}
\label{sec:scope}

Sec.~\ref{sec:hardware} describes what each design choice bought. We
set out here what each one cost, so a prospective user can judge fit
before purchase.

\emph{Quantitative characterization.} This revision reports the design
and its rationale rather than measured performance. Repeatability
under arm moment load, payload sag, end-effector precision, grip-force
traces, and protection-event statistics are the figures that would let
a reader check these claims without buying the robot, and they are the
first thing we intend to add.

\emph{Modification.} We do not open-source the robot as a whole.
Mechanical modification is supported where labs most often need it:
end effectors can be designed against the open connector interface,
and we supply replacement CAD for parts that break. We stop short of the package needed to re-engineer the platform from
scratch, which is the case for a group whose research question is the
design of the robot itself. Such a group is better served by an open
platform. For everyone whose question is what to do with a robot, the
distinction rarely arises in practice.

\emph{Payload.} 1.5\,kg per arm suits domestic manipulation and is
below the platforms above the \$5{,}000 tier in
Table~\ref{tab:cost}.

\emph{Base mobility and speed.} Lateral repositioning is a compound
maneuver rather than a strafe, in exchange for a drive costing a
fraction of the swerve modules that account for \$2{,}700 of YOR's
bill of materials. The column traverses its 76\,cm in about 25
seconds, which domestic tasks tolerate easily and research on dynamic
manipulation would not.

\emph{Compliance.} The arms are position-controlled rather than
running a joint-stiffness controller with gravity compensation.
Compliance is local to the contact point, supplied by the TPU fingers,
and Sec.~\ref{sec:protection} handles unexpected contact by detecting
and backing off rather than by conforming.

\emph{Inference location.} Running inference off-board keeps per-unit
compute at the cost of a single-board computer, at the price of a
network dependency.

\section{Conclusion}

Nori A3 sells assembled for \$1{,}688, below the parts cost of the
cheapest comparable research platform. Reliable actuation, useful
payload, floor-to-counter reach, and grip-force feedback are usually
assumed to be out of reach at that price. We recover them with
software running on the components the price allows: a protection
architecture that removes the dominant failure mode of commodity
serial-bus servos, and a sensorless force channel that turns a current
register into an observation. The protection layer is released for the
actuator class it targets. What this paper sets up without closing is
the argument that unit cost bounds the rate at which matched
demonstration data can be collected, which we intend to report on
next.

\section*{Acknowledgements}

The author thanks Sungjoon Park and Wen Ni Chew, co-authors on the
prior version of this platform~\cite{li2026noribot}, whose work on the
lift, the actuator protection stack, and the sensorless force channel
this paper builds upon; Alexander Roller, Michael Ma, and Shanying Liu
for their ongoing work on the platform; and the wider Nori Robotics
team for the design and manufacture of A3.

\bibliographystyle{IEEEtran}

\end{document}